# International Conference on New Trends in Engineering & Technology (ICNTET)

## 7th & 8th September 2018

## ICNTET Conference Proceedings

*Organised by*

## GRT Institute of Engineering and Technology

Tiruttani, Thriuvallur, Chennai, Tamilnadu-India.

ISBN : CFP18P34-PRT/978-1-5386-5629-7

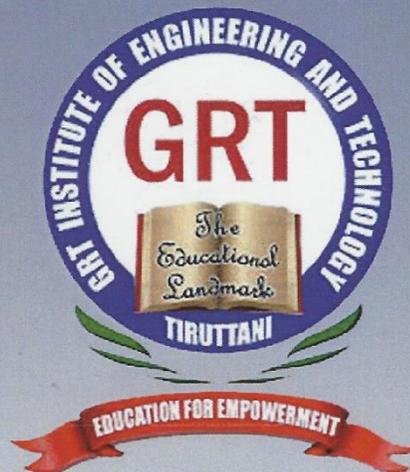

**Supported by**

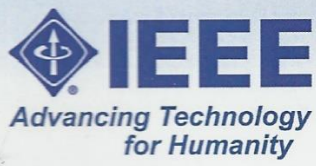 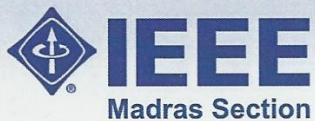 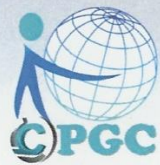 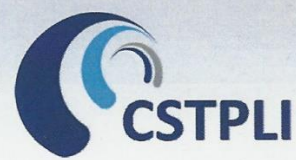

maintaining the temperature while taking the FTIR spectra. To remove the noise, Savitzky Golay smoothing method is used. The main aim of a deconvolution technique is to remove the instrumental widening in peaks so as to produce the pure spectrum, without any instrumental distortion. Errors due to shadowing and sunlight can be neglected. In general, most of the spectral instrument response functions are Gaussian in form. The widened convoluted peak consists of mainly two function, one is natural function and other one is Gaussian function. These spectral peaks are usually formed by a natural Lorentzian widening and an independent Gaussian instrumental widening. The convolution of a Gaussian and Lorentzian function is an another function, which can also be interpreted from the de-convolved Spectra. This study shows an approach for spectral deconvolution that accurately represents absorption bands as discrete mathematical distributions and resolves composite absorption features into individual absorptions bands. The FTIR Spectra of basic igneous(basalt) is taken at a temperature of 60° Celsius (Figure-1), maintained using a manually controlled oven. The hidden peaks obtained after baseline correction and peak fitting with deconvolution for basic igneous is shown in Figure-2. From this study it is concluded that the various regions of different bands of the spectrum, are Gaussian in nature. The baseline correction method is used after deconvolution, so that the result of peak fitting is assured with accurate resulting parameters. Following peaks are observed after deconvolution: 9.03μm, 9.2μm, 9.47μm, 9.79μm, 10.13μm, 10.25μm, 10.9μm, 11.03 μm. It is evident that basalt shows its absorption features around 9.7μm and there is also a 9-11μm-Si-O-Si asymmetric stretch (Mitra, 1996). As the deconvoluted peaks observed between 9-11μm shows asymmetric variations, it can be concluded that Si-O-Si stretch is there in the range of 9-11μm and their accurate band values can be located easily. Initially the peaks are broad and are asymmetric in nature but the bands obtained after deconvolution revealed to be single peak bands and are symmetrical(Kim et al., 2017). The band locations 11.03,11.57,11.93,12.36,13.06,13.4μm shows the occurrence of carbonates and nitrates. Absorptions at rock surfaces consisting mixed minerals are complex and only partially understood. This study can give a significant way to evaluate compositional patterns from laboratory data and in recognizing mineral constituents in remotely sensed spectra. This algorithm can effectively extract the required parameters for identification of absorption features in a rock spectrum.

## *Visual Navigation for Airborne Control of Ground Robots from Tethered Platform Creation of the First Prototype*

*Ilan Ehrenfeld, Max Kogan, Oleg Kupervasser, Vitalii Sarychev, Irina Volinsky, Roman Yavich and Bar Zangbi*

Abstract: We propose control systems for the coordination of the ground robots. We develop robot efficient coordination using the devices located on towers or a tethered aerial apparatus tracing the robots on controlled area and supervising their environment including natural and artificial markings. The simple prototype of such a system was created in the Laboratory of Applied Mathematics of Ariel University (under the supervision of Prof. Domoshnitsky Alexander) in collaboration with company TRANSIST VIDEO LLC (Skolkovo, Moscow). We plan to create much more complicated prototype using Kamin grant (Israel)

## *Neural Networks Based Obstacle Avoidance and Path Planning for a Mobile Robot through Image Processing*

*Ragul, Renjith Ganesh and Nithya. M*

Abstract: Path planning and obstacle avoidance are inevitable subset of robotics. In this work, an image of the environment where the mobile robot is to be deployed is given as the input of the system. The image may consist of multiple obstacles and target. An object detector module is used to detect the obstacles in the image. An algorithm based on Pulse Coupled Neural Network (PCNN) method is

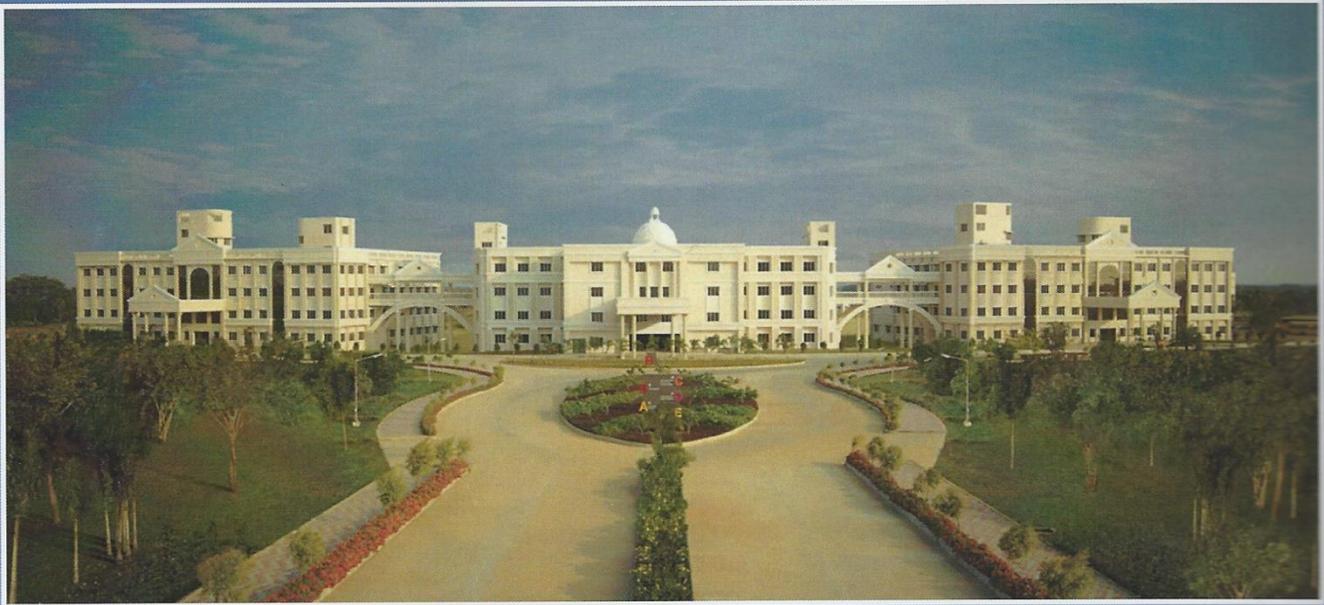

**Supported by**

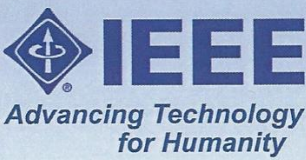 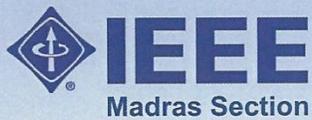 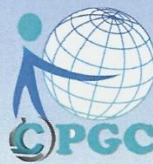 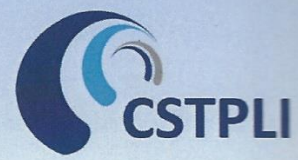

Website : www.icntet.org | www.pgcinfo.org
Contact : 0891-2761433

# *Visual navigation for airborne control of ground robots from tethered platform*

*creation of the first prototype*

Ilan Ehrenfeld[1], Max Kogan[1], Oleg Kupervasser[1,2], Vitalii Sarychev[2], Irina Volinsky[1], Roman Yavich[1], Bar Zangbi[1]

1. Ariel University, Ariel, Izrael
2. TRANSIST VIDEO LLC, Moscow, Skolkovo, Russia
olegkup@yahoo.com

**Abstract** -We propose control systems for the coordination of the ground robots. We develop robot efficient coordination using the devices located on towers or a tethered aerial apparatus tracing the robots on controlled area and supervising their environment including natural and artificial markings. The simple prototype of such a system was created in the Laboratory of Applied Mathematics of Ariel University (under the supervision of Prof. Domoshnitsky Alexander) in collaboration with company TRANSIST VIDEO LLC (Skolkovo, Moscow). We plan to create much more complicated prototype using Kamin grant (Israel)

1. Introduction

The system relates to the control systems of automated devices and may be used for the coordination of the ground movable automated devices (automated transport, automated agricultural machines, municipal and aerodrome vehicles, garden lawnmowers and so on), hereinafter referred to as the robots.

The invention essence is the system of navigation [1-20] and intercoordination of one or more roots located on the controlled area including one or more robot tracing devices on the suspended platforms, natural or artificial markings, central module for robot coordination and orientation detection to which the information is transferred from all the tracing devices, charger and the system is equipped with the central calculation module located on the suspended platform, on the ground, on the charger or on the robot designed with the possibility to detect the coordinates, to orient the system and to form the control commands based on the information received from all the above described devices.

The technical result is the development of the robot efficient coordination using the devices located on the towers or the aerial apparatuses tracing the robots on the controlled area and supervising their environment including natural and artificial markings. One of the developments is to take into account delays in the control system. The technical result coincides with the engineering challenge.

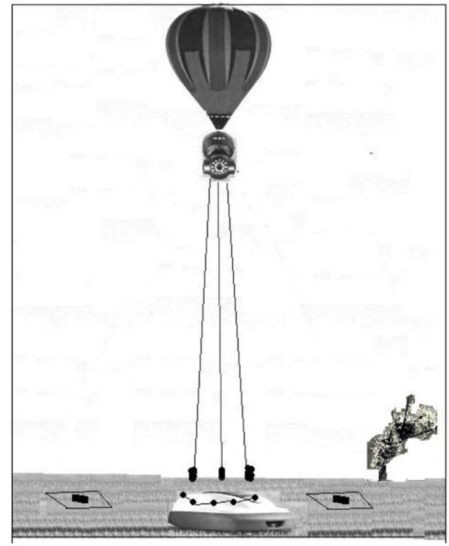

FIG. 1

The system relates to the control systems of automated devices and may be used for the coordination of the ground movable automated devices (automated transport, automated agricultural machines, municipal and aerodrome vehicles, garden lawnmowers and so on), hereinafter referred to as the robots.

2. Main Ideas

*A. Engineering Challenge*

The engineering challenge to be solved by this invention is the development of the robot efficient coordination using the devices located on the towers or the aerial apparatuses tracing the robots on the controlled area and supervising their environment including natural and artificial markings. The technical result coincides with the engineering challenge.

*B. Background*

One of the major challenges of navigation, coordination and control of robots is the absence of cheap and reliable system of navigation and situational action coordination. For instance, in order that the robot-lawnmower does not leave the mowing area it must be marked by the wire, they have random navigation methodologies, they are inconvenient, static, expensive technology. See the Internet publication dated June 15, 2012: [1]

Recently the systems of IR fences or markings are offered. Also, it is possible to use the ground radio beacons. But this approach complicates the system substantially.
Using of GPS and even more precise DGPS systems implies a number of problems:

- GPS signal near houses may be shielded, re-reflected or just damped by interferences or purposely resulting in the robot loss of coordination;
- it is required to measure the coordinates of the operation area limits (for instance, the operation area for the robot-lawnmower) and to point them out to the robot that is a labor-intensive process;
- these systems give coordinates but not the robot orientation;
- the robots orientate by abstract coordinates not by the real world (so if there is a stationary or moving obstacle (a dog or a child) then the system will not detect it/him);
- these systems may not detect where on the area there is grass not mowed;
- only with DGPS or GPS it is hard to organize the intercoordination of the robots that are not aware of the interposition and must have the complex system of mutual detection and signal exchange;
- the satellite systems have high working cost.

Many of these problems may be solved by a video navigator set on the robot. But such solution has a limited visibility area that may be expanded only by using a large number of cameras with a wide visibility area that considerably complicates the system.

Besides a number of well detectable ground markings must be used. Natural landmarks do not always have these properties, so the mowing area must be marked by ground markings.

The intercoordination of the robots keeps being a sophisticated way including using of a complex system of the artificial vision and decentralized object detection system. The decentralized control system of joint activity is multiply more complex and more expensive than one centralized system.

The engineering challenge to be solved by this invention is the development of the robot efficient coordination using the devices located on the towers or the tethered aerial apparatuses tracking the robots on the controlled area and supervising their environment including natural and artificial markings. One of the developments is to take into account the delay in the control system and using this delay in stabilization of the motion. The technical result coincides with the engineering challenge.

*C. Summary of Technology*

To support the challenge it is supposed the system of navigation and intercoordination of one or more roots located on the controlled area including one or more robot tracing devices on the suspended platforms, natural or artificial markings, central module for robot coordination and orientation detection to which the information is transferred from all the tracing devices, ground charger and the system is equipped with the central calculation module located on the suspended platform, on the ground, on the charger or on the robot designed with the possibility to detect the coordinates, to orient the system and to form the control commands based on the information received from all the above described devices.

And the rotor-driven aerial apparatus may be connected to the charger or the robot with the possibility to receive/ transfer the energy. The system may be additionally equipped with one or more upper hemisphere tracing devices located on the surface of the controlled area or on the robots designed with the possibility to receive/ transfer the information with the central calculation module and one or more device for sunlight-to-electricity conversion located on the rotor aerial apparatus and/ or robot and/ or on the surface of the controlled area.
The sunlight-to-electricity conversion device may be designed with the possibility to transfer the energy to one or more chargers for battery charging of at least one robot and/ or at least one rotor-driven aerial apparatus in the helicopter mode. One or more rotor-driven aerial apparatus may be designed with the possibility to transfer the energy output in the wind-driven motor mode to the energy storage device located on the rotor-driven aerial apparatus and/ or one or mode charger to be used in the helicopter mode. The rotor device may be used as the drone-guard for a house or an area

Unlike the systems used GPS the tracing devices (one or more cameras) are located over the controlled area before the robot operation and the place and the height depend on the visibility conditions of the controlled area. That is compared to the GPS systems located whatever the robot coordination purposes the tracing devices are located exactly for convenient coordination of the robots solving the GPS problems, for instance, shielding and re-reflection of the signal. It is to be noted that the GPS satellites are not the environment or the robot tracing devices which coordinates must be detected. Vice versa the robots themselves are the GPS satellite tracking systems and the robot coordinates may be detected by GPS only on the robot itself and provided that there are three or more GPS system satellites in its available space environment.

The efficiency of detecting the coordinates (space or angular) and the intercoordination of the robots is reached due to the devices for coordinating the robots including aerial apparatus or those located on the tower tracing the robots on the controlled area and supervising their environment including natural and artificial markings where the module for detecting the aerial apparatus coordinates designed with the possibility of the information exchange with another module located on the same devices with the possibility to detect the coordinates of the controlled robot. And the mentioned module is designed with the possibility to receive and transfer the control commands and signals to the controlled robot. The aerial device located above, or the device located on the tower may be:

1. UAV,
2. tower-antenna,
3. high-altitude tethered platform of continuous supervision (tethered aerostatic airships or constant-level balloons),
4. tethered aerodynamic rotor-driven due to the electric energy supplied to the rotors (similar to tethered helicopter platforms of Hovermast-100 by Skysapience),
5. Tethered rotor-driven aerial apparatuses with aerodynamic unloading due to the energy of the high-altitude wind (autorotation) always existing at high altitude (approx. 4 m/s at 100 m, Fig. 2), for instance, tethered autogyros and gyroplanes (similar to tethered autogyros Fa330 used by the Germans during the World War II).

However, each of these methods has drawbacks:

1. UAV are expensive and complicated for controlling and calculation, have limited continuous observation time,
2. the towers are hard to install and locate or re-install,
3. the tethered aerostatic airships or constant-level balloons require a compound pumping mechanism and are inconvenient for stabilization,
4. tethered aerodynamic rotor-driven vehicles required much energy,
5. the tethered autogyros and gyroplanes do not fly if no wind.

To prevent these drawbacks, we can use some combination of such devices.

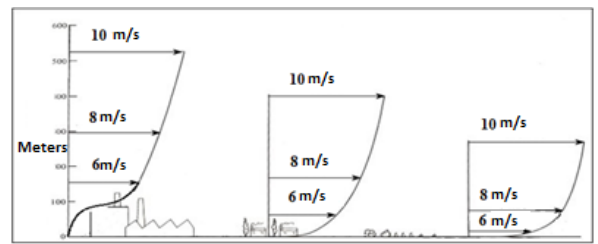

Fig. 2

*D. Engineering Challenge*

The engineering challenge to be solved by this invention is the development of the robot efficient coordination using the devices located on the towers or the aerial apparatuses tracing the robots on the controlled area and supervising their environment including natural and artificial markings. The technical result coincides with the engineering challenge.

*E. Summary of Invention*

To support the challenge it is supposed the system of navigation and intercoordination of one or more roots located on the controlled area including one or more robot tracing devices on the suspended platforms, natural or artificial markings, central module for robot coordination and orientation detection to which the information is transferred from all the tracing devices characterized in that at least one suspended platform is a rotor device able to operate in the following modes:

a) autogyro due to the relative airflow,
b) wind-driven motor receiving the energy from the ram wind,
c) helicopter receiving the energy from the ground charger and the system is equipped with the central calculation module located

on the suspended platform, on the ground, on the charger or on the robot designed with the possibility to detect the coordinates, to orient the system and to form the control commands based on the information received from all the above described devices.

And the rotor-driven aerial apparatus may be connected to the charger or the robot with the possibility to receive/ transfer the energy. The system may be additionally equipped with one or more upper hemisphere tracing devices located on the surface of the controlled area or on the robots designed with the possibility to receive/ transfer the information with the central calculation module and one or more device for sunlight-to-electricity conversion located on the rotor aerial apparatus and/ or robot and/ or on the surface of the controlled area. The sunlight-to-electricity conversion device may be designed with the possibility to transfer the energy to one or more chargers for battery charging of at least one robot and/ or at least one rotor-driven aerial apparatus in the helicopter mode. One or more rotor-driven aerial apparatus may be designed with the possibility to transfer the energy output in the wind-driven motor mode to the energy storage device located on the rotor-driven aerial apparatus and/ or one or mode charger to be used in the helicopter mode.
The rotor device may be used as the drone-guard for a house or an area.

*F. Brief Description of Drawings*

The Fig. 2 shows the wind force-height dependency diagram for different types of area: urban, village and settlement.

The Fig. 3 shows the variants of the project realization. And the following designations are used: tethered unmanned aerial apparatus (tethered UAV) 1, charging and control unit 2, with the camera (cameras) 3, markings on the ground 4 and on the robot 5, natural markings - bush 6.

*G. Embodiment of Invention*

During the embodiment of the invention the centralized robot control system is created and the accuracy of their coordinate (space or angular) detection is increased.
Using of tethered platforms with the supervision units that may operate in three different modes: autogyro, wind-driven motor and helicopter provides the efficient robot coordination using the devices located on the towers or the aerial apparatuses tracing the robots on the controlled area and supervising their environment including natural and artificial markings.

The joint use of different modes enables to add each and to compensate the drawbacks of each separate. The solution is explained on the fig 3 where three possible variants (a, b and c) of the supposed system realization are shown. The fixed cameras covering the whole lower hemisphere are on the suspended platform. It is cheaper than the one controlled camera and the wire communication channel (optic fibre or twisted pair) is reliable and intensive. Several cameras may be located on the tethered platform and on its gimbal - at the required small height. The tethered platforms are fixed to the ground by the rope (fig. 3a) or to the energy accumulation and output unit by the cable (fig. 3b) or directly to one of the robots on the controlled area (fig. 3c).
Also, the energy may be output due to the solar batteries mounted on the tethered platform, ground or the robots.
The supposed solution may use the relative (differential) robot video positioning both regarding the area and aerial apparatus (tower). The robot coordination from the aerial apparatus (tower) may not always require the coordinates of the UAV supervision itself. Precise relative robot positioning regarding 3 and more special markings, fixed ground objects and other ground robots is possible.
Precise UAV coordinates do not ensure precise ground robot coordinates. However, these UAV coordinates (position and orientation) may be needed for correcting the project distortions of the received images.
Passive video supervision in broad and artificial daylight is possible. The all-weather capability is provided by the IR and radio locating vision, passive reflectors and active IR markings, IR LEDs and so on.
Using of several supervision cameras over the controlled area (different combination of on fixed and high-altitude tethered platforms) improves the reliability, stereoscope positioning accuracy, eliminates the dead areas (for instance, behind and under the trees).
The markings easy to be seen from above may be installed on the robot and on its charger.
The central module where all the information from all the tracing devices is transferred detects the coordinates and the orientation of the one or more controlled robots (relative (differential) robot video positioning both regarding the area and the tracing devices (cameras) and if necessary detects the coordinates and the orientation of the tracing devices. And the mentioned module is designed with the possibility to transfer the control commands and signals (including RF) to the robots, tracing devices, chargers and the exchange of control and information signal between them is possible.
If there are several robots, then their centralized coordination is simple - the cameras shows them all simultaneously and the unified computer system coordinates their joint movement. The boundaries of the operation area (for instance, mowing for the robot lawnmower) may be set by setting the boundaries on the screen of the computer system by the area image (for instance, using the mouse arrows or drawing by the sensor pen or the finger on the screen).

The developed system functions as follows: firstly, one or more robots are located on the controlled area (for instance, a lawn). Before the robot operation the tracing devices (one or more cameras) are located on the aerial apparatus or the towers over the controlled area and the places and the height of the gimbal depend on the visibility conditions of the controlled area.

It is also possible that starting the operation the tracing device is located on the ground or on one or more robots and the while operating it may fly up, fly or land on the robot tracing towers on the controlled area.

The tethered platform tracing devices may be located on the ground and on the robots enabling to detect the interposition and orientation of the tracing devices and the robots and to detect the robot rotation angle more accurately and to find the robot position in the camera dead areas (under the shelters or under the trees) by orientation by the shelter ceilings or the tree leaves above the robot.

Besides instead of the visible signal other spectra sections may be used. And the signal may be not only natural but generated by the robot or the device on the camera or in other space point. Sound, ultrasound signals, radio detection, sensors and markings, for instance, odour or chemosignals or radioactivity a little bit above the background level (for instance, silicon slices) may be used.

The supervision system is capable to detect the obstacles or the moving objects, determines the level and the quality of grass mowing. It is easy to realize and has low cost price.

This system may be used for a wide class of robots: automated lawnmowers, robots for cleaning the rooms, tractors, snow-removal, garbage disposal and flushing vehicles, vehicles for people and goods transportation, agricultural and municipal vehicles, transport and so on. This system may be used for extra-terrestrial robots on other planets, for instance, for Mars rovers.

The system easily stays within the frames of the "smart home" or even "smart city" enabling to coordinate simultaneously a lot of actions, robots and other control objects and to solve many tasks - for instance, not only navigation but detection.

The invention is disclosed above with reference to the concrete variant of its implementation. The experts may know other variants of the invention embodiment not altering its essence as it is disclosed herein. Thus, the invention description is deemed to be limited by volume only by the invention formula below.

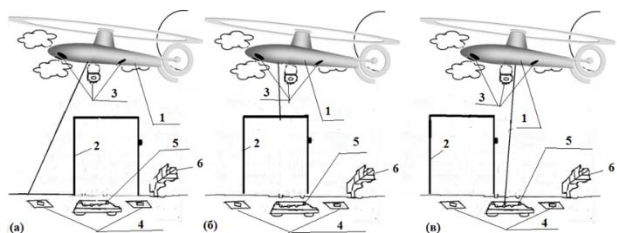

Fig. 3

3. Simple prototype creation

A. Purpose of the prototype creation.

We would like to develop
1) a simple <u>physical</u> model (a prototype) that can demonstrate possibility of visual navigation for airborne ground robots control
2) patents for creating IP

The project includes 3 components:
a) computer with controlling program
b) moving ground robot-toy. The robot connected to controlling computer by the help of wireless connection (for example, Wi-Fi, Bluetooth, RW). Computer sends a controlling signal using this wireless connection
c) video-camera connected to ceiling. The camera connected to the controlling computer by the help of video-wire. The computer program uses the video-information from the camera for measurement a deviation from desirable way and forms controlling wireless signal to the robot for the movement correction.

The prototype demonstrates in physical reality possibility of visual navigation and controlling the robot: 1) the robot moves according defined path 2) the robot moves along a random way, but in some pre-defined boundaries.

B. Results the prototype creation.

We developed the prototype and got the desirable 2 types of motion.
We send 2 films that demonstrate these 2 types of motion.
On the films we can see screen of the controlling computer. On the screen we see video from the camera and computer marks on the video (red desirable trajectory or red desirable boundaries, green rectangle around robot-toy (tank)).
Initially, we draw the red desirable trajectory for the first video or the red desirable boundaries for the second video.

On the second part of the video, we can see that computer program can find the robot-toy (tank) on the video and can track the tank. Indeed, we can see green rectangle around the tank during all path on the video.

On the first video (fig 4.), we can see that the tank follows the desirable trajectory (in spate of sometimes appearing deviations from the desirable trajectory). The system completely automatically compensates these deviations.

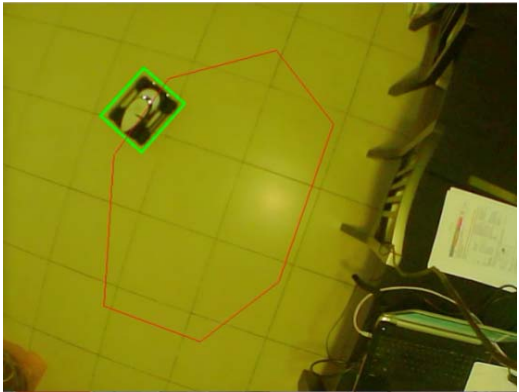

Fig. 4

On the second video (Fig. 5), we can see that the tank follows random way, but reflects out from desirable boundaries. The system completely automatically revers the tank motion near the boundaries.

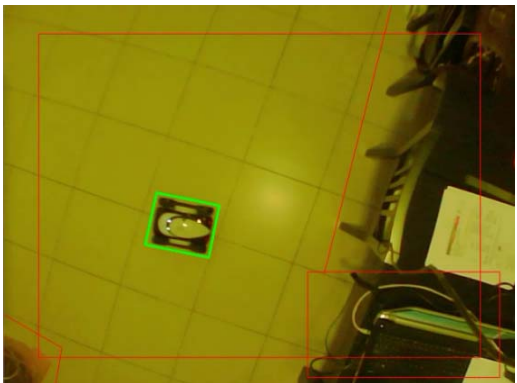

Fig. 5

4. Future plans and conclusion

We plan to create much more complicated prototype using Kamin grant (Israel).
The plane consists of the next stages:

1. Creating programs of video-navigation for airborne control from towers of ground robots on flat ground surface
2. Design and integration of robotic system: observation towers, controlling center, and ground robots
3. Testing robotic system and computer program, errors correction of robotic system and computer program
4. Rewriting programs for video-navigation than airborne control is made from tethered platform (drone or balloon) and ground robots on not-flat surface
5. Modernization of robotic system: creation of observation tethered platform (drone or balloon) and modernization of ground robots for not-flat surface
6. Testing robotic system and computer program, errors correction of robotic system and computer program

This system may be used for a wide class of robots: automated lawnmowers, robots for cleaning the rooms, tractors, snow-removal, garbage disposal and flushing vehicles, vehicles for people and goods transportation, agricultural and municipal vehicles, transport and so on. This system may be used for extra-terrestrial robots on other planets, for instance, for Mars rovers.

The system easily stays within the frames of the "smart home" or even "smart city" enabling to coordinate simultaneously a lot of actions, robots and other control objects and to solve many tasks - for instance, not only navigation but detection.